%% file: sample-sigconf.tex
\documentclass[sigconf]{acmart}
\AtBeginDocument{%
  }

\copyrightyear{2026}
\acmYear{2026}
\setcopyright{cc}
\setcctype{by}
\acmConference[DAC '26]{63rd ACM/IEEE Design Automation Conference}{July 26--29, 2026}{Long Beach, CA, USA}
\acmBooktitle{63rd ACM/IEEE Design Automation Conference (DAC '26), July 26--29, 2026, Long Beach, CA, USA}
\acmDOI{10.1145/3770743.3804046}
\acmISBN{979-8-4007-2254-7/2026/07}

\input{packages}
\begin{document}

\title[\dd: Segmenting Patterns from Defects in Wafer Manufacturing Using Weak Supervision]{\dd: Segmenting Patterns from Defects in Wafer Manufacturing Using Weak Supervision}

\author{Dain Kwon}
\affiliation{%
  \institution{Seoul National University}
  \city{Seoul}
  \country{South Korea}
}
\email{dain.kwon@snu.ac.kr}

\author{Changmin Shin}
\affiliation{%
  \institution{Seoul National University}
  \city{Seoul}
  \country{South Korea}
}
\email{scm8432@snu.ac.kr}

\author{Sunjong Park}
\affiliation{%
  \institution{Seoul National University}
  \city{Seoul}
  \country{South Korea}
}
\email{ryan0507@snu.ac.kr}

\author{Kanghyun Choi}
\affiliation{%
  \institution{Seoul National University}
  \city{Seoul}
  \country{South Korea}
}
\email{kanghyun.choi@snu.ac.kr}

\author{Hyeyoon Lee}
\affiliation{%
  \institution{Seoul National University}
  \city{Seoul}
  \country{South Korea}
}
\email{hylee817@snu.ac.kr}

\author{Jaewon Jang}
\affiliation{%
  \institution{SK Hynix}
  \city{Seoul}
  \country{South Korea}
}
\email{jaewon2.jang@sk.com}

\author{Minseok Choi}
\affiliation{%
  \institution{SK Hynix}
  \city{Seoul}
  \country{South Korea}
}
\email{choineral@gmail.com}

\author{Jinho Lee}
\orcid{0000-0003-4010-6611}
\affiliation{%
  \institution{Seoul National University}
  \city{Seoul}
  \country{South Korea}
}
\email{leejinho@snu.ac.kr}

\renewcommand{\shortauthors}{Dain Kwon et al.}

\begin{abstract}
In semiconductor manufacturing, defect analysis is essential, but manual inspection cannot scale.
However, existing automated inspection methods %
remain insufficient
for root-cause analysis and process optimization.
To this end, we present \dd, a weakly supervised %
wafer %
defect segmentation 
method. %
\dd enables %
pixel-level separation of patterns by leveraging only image-level annotations.
It consists of a three-phase %
training: encoder pretraining, %
knowledge transfer to learn spatial cues, %
and training on synthetic mixed-defect data for accurate %
segmentation. %
Experiments 
demonstrate that \dd %
outperforms the baselines.
The code is available at \url{https://github.com/meowrowan/SePArate}.
\end{abstract}


\maketitle

\section{Introduction}

\begin{figure}[t]
  \centering
  \includegraphics[width=0.92\linewidth]{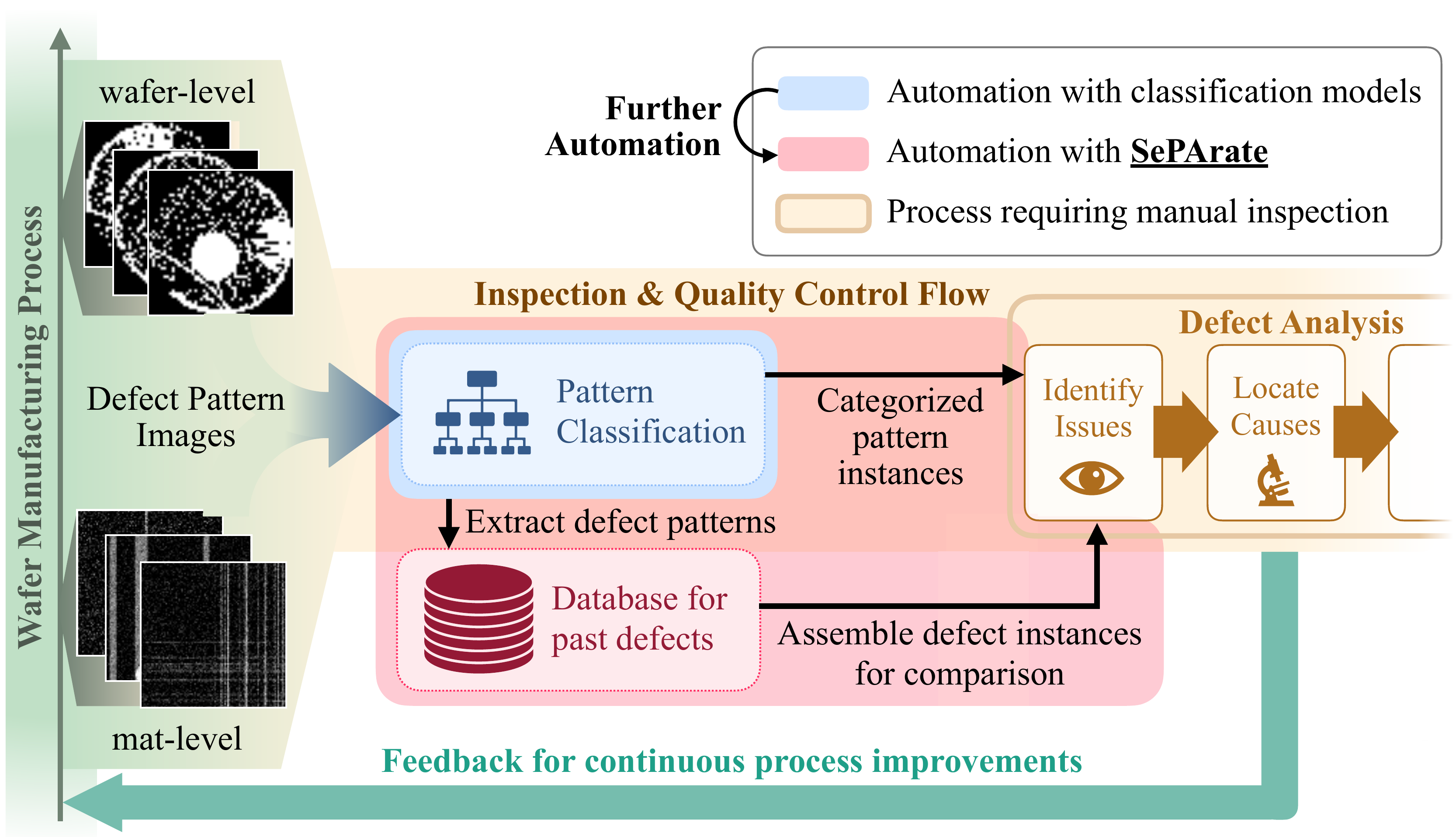}
  \caption{Automation of the overall visual inspection of defects and continuous improvement by integrating 
\dd.}
    \label{fig:automation_scenario}
\end{figure}

Defect %
analysis is an essential step in semiconductor manufacturing to minimize faults during production. 
By inspecting %
patterns of failures, engineers can trace underlying causes of defects, leading to process refinement based on the found issues~\cite{gleason1998rapid,shankar2005defect,getachew2024data}.
\fig{fig:automation_scenario} illustrates an inspection scenario. 
During the production, defects are detected, either at the wafer-level, or at the specific component-level (e.g., mat-level for a DRAM~\cite{isscc_ddr4}).
These defect maps become classified images organized into datasets. Then, experts analyze these to identify issues, trace their causes, and provide feedback.

However, the sheer volume of inspection data makes manual inspection infeasible. 
Modern manufacturing processes generate terabytes of defect data each day~\cite{waferdata1,waferdata2}, containing diverse and intricate pattern variations that could guide further manufacturing process optimization. 
In such conditions, human inspection cannot scale to the full dataset.
Therefore, automated defect pattern analysis is a natural progression for wafer defect detection.

Despite recent advances, current AI-based defect analysis methods~\cite{palm, waferssm} remain limited in practical utility. 
Many of them are classification methods, i.e., they detect whether predefined patterns exist on wafers. 
This leads to a significant limitation: 
\emph{They discard geometric attributes such as shape, area, and orientation. }
These attributes provide essential information about the root cause of the defects and clues on how to fix them.
Consequently, automated pixel-level defect pattern annotation is necessary, but AI-based segmentation methods do not support wafer defect analysis and are hard to train due to the absence of a large, annotated dataset.

To address this, we propose \dd: a weakly supervised defect pattern segmentation approach requiring only classification labels without costly pixel-level annotations. 
\dd generates pixel-level masks that isolate individual patterns from defect images, enabling finer categorization, new pattern discovery, and automated database construction that previously required manual curation.
An end-to-end automation of the defect analysis pipeline can be initiated, as illustrated in \fig{fig:automation_scenario}. 
Using a combination of %
self-generated supervision and a task-specific fine-tuning approach, \dd effectively isolates diverse defect patterns. 
Results on a production-line DRAM mat-level dataset and a public wafer-level dataset demonstrate that \dd significantly outperforms baselines. The key contributions of our work are as follows.

\begin{itemize}[leftmargin=*, itemindent=.8em]
\item We are the first to tackle \textbf{segmentation for binary mixed-type defect pattern images} with weak supervision.
\item We devise \textbf{\dd}, a framework 
that trains a defect pattern segmentation network 
under image-level supervision, without requiring additional costly instance annotations.
\item %
\dd greatly surpasses existing baselines, %
highlighting the value of its defect-aware design. 
\end{itemize}

\section{Background}
\label{sec:background}

\subsection{Defect Data in Wafer Manufacturing}
Defects
occur at 
various granularities in wafer manufacturing, as shown in the leftmost side of \fig{fig:automation_scenario}. 
For generalization, 
we use two datasets of different granularity:
a private, DRAM mat-level defect dataset \mtdata, obtained from an actual DRAM manufacturing process, and a public wafer-level defect dataset \wmdata\ \cite{mixedwm38}. 
The pixels are binary: 0 for normal, and 1 for defects.

\begin{figure}[t]
\centering
\includegraphics[width=\linewidth]{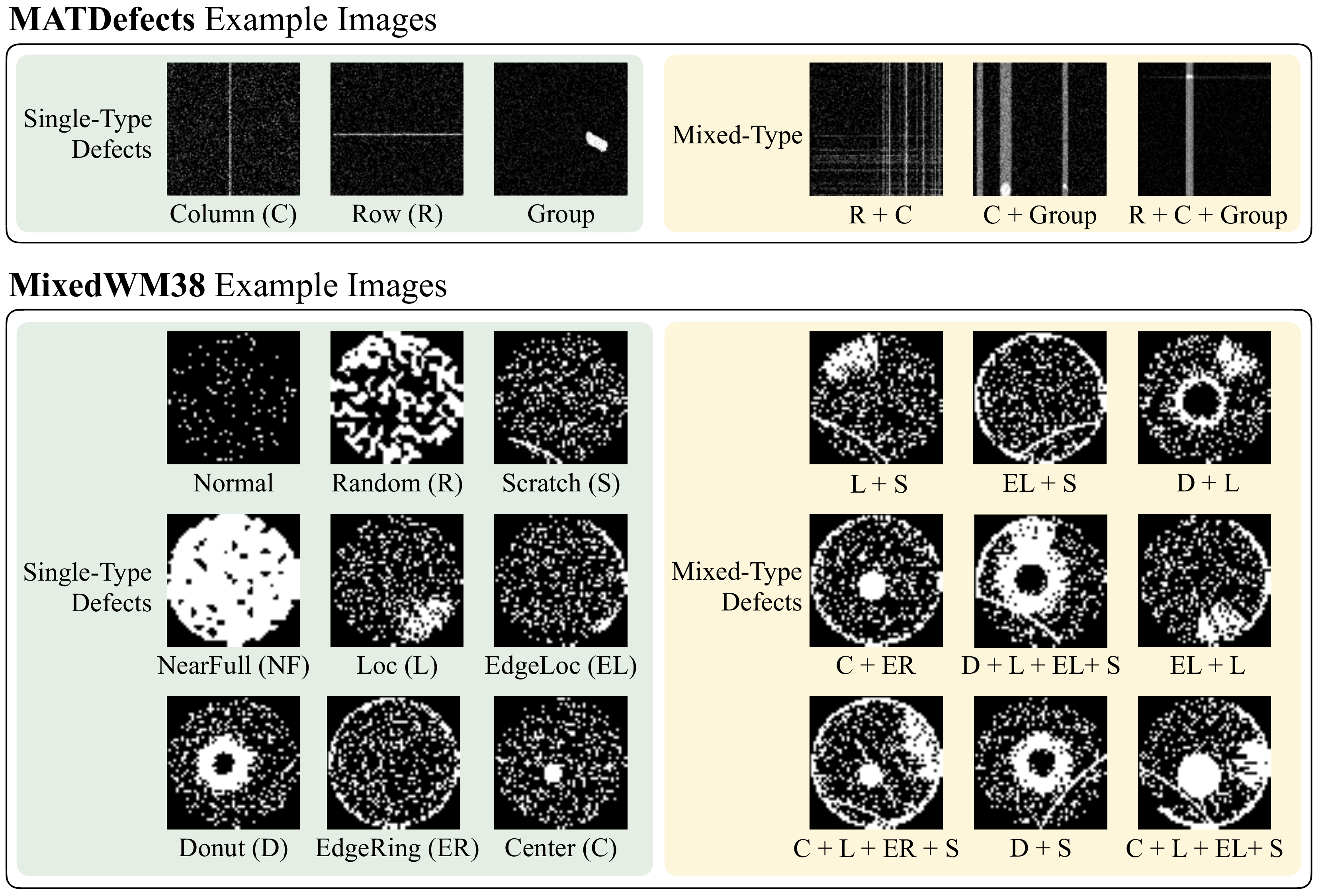}
\caption{Single and mixed-type defect pattern examples for %
\mtdata\ and \wmdata. 
}
\label{fig:dataset}
\end{figure}

\mtdata 
contains data %
from DRAM products, where each pixel represents a DRAM cell (i.e., a bit).  
A DRAM die~\cite{isscc_ddr4} often exhibits a regular structure composed of multiple banks, %
each comprising multiple mats. %
Each mat is a square cell array, typically 512-2,048 in size, depending on the product. 
\mtdata includes 
single and mixed-type defect patterns.
The single defect patterns are classified as: 
row-wise (wordline or wordline driver issues), column-wise (bitline or sense amplifier issues), or group (individual cell issues). 
As %
in \fig{fig:dataset}, multiple single-type defects may appear in an image, forming a mixed-type defect pattern.

\wmdata contains images generated from full wafers, where each pixel corresponds to a die. 
It contains %
38 single and mixed-type patterns,
with 8 single-type patterns, also shown in \fig{fig:dataset}. 

\begin{figure*}[ht]
    \centering
    \includegraphics[width=0.95\textwidth]{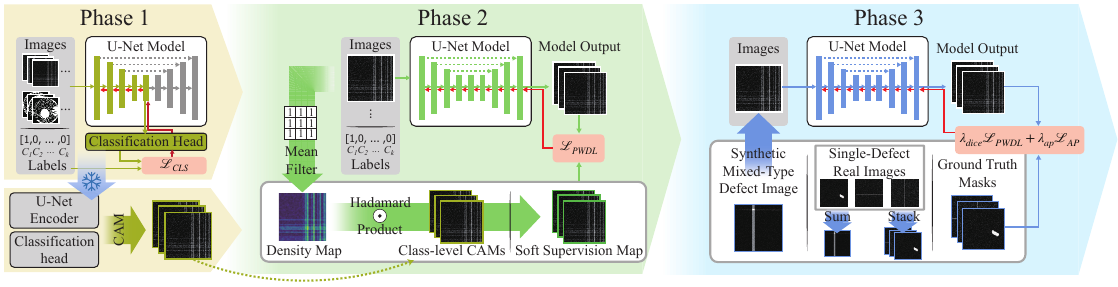}
    \caption{A full overview of our proposed method, \dd. 
    i) Phase 1 trains the encoder for multi-label classification, enabling the model to learn about defect patterns using image-level labels.
ii) Phase 2 transfers classification knowledge to the U-Net by training with soft supervision maps and pattern-weighted Dice loss.
iii) Phase 3 further tunes the model using synthetic mixed-defect data, improving precision with pattern-weighted Dice loss and absence-penalty scheduling.  
    }
    \label{fig:method}
\end{figure*}

\subsection{Segmentation without Ground-Truth Data}
\label{sec:wsss}

Defect pattern datasets usually provide only image-level ground-truths, but we need pixel-level annotation for each pattern to do segmentation.
Two approaches are feasible: 
zero-shot foundation models for segmentation \cite{clipseg, groundedsam}, or \emph{weakly supervised semantic segmentation (WSSS)} methods \cite{mctformer, recam, sam2cams}. %

Zero-shot models such as CLIPSeg~\cite{clipseg} and GroundedSAM~\cite{groundedsam} 
use vision-language pretraining to align visual features with text semantics. 
As they %
enable text-guided segmentation without labeled examples, defect-pattern text prompts may be used to guide them.

There are also WSSS methods \cite{mctformer, recam, sam2cams}, which train segmentation models with image-level labels.
Most methods train a classifier, generate pixel-level pseudo-labels from intermediate activations (e.g., Class Activation Maps (CAMs) \cite{recam, sam2cams} or attention maps \cite{mctformer}), then train a segmentation model with these synthetic labels.

The main challenge with these alternatives is that they rely on visual cues--such as color, texture, or shading--that are largely absent in our binary data. 
With no diversity in pixel values to infer about shape, size, or depth, we are only left with geometric structure to differentiate between patterns.

Therefore, a new defect-aware approach is required: a weakly supervised framework that explicitly leverages structural and distributional cues without depending on visual semantics. %
\dd is explicitly designed for this setting, %
providing a practical solution for defect pattern segmentation.

\subsection{Existing Works for Binary Defect Data} %
For \emph{binary defect images}, existing works have %
focused on 
multi-label classification, detecting the presence of each
single-type defect pattern in mixed-type defects.
The classification models evolve from 
CNNs \cite{mixedwm38, kyeong2018classification,kim2023advances,yu2019enhanced,shon2021unsupervised,saqlain2020deep} to transformers and state space models \cite{palm, waferssm}. 
There exist a few segmentation works, \cite{wafersegnet, nakazawa2019anomaly} %
 but they assume only a single defect instance per image and thus cannot separate multiple single-type defect patterns.

One work, SSB-Rec \cite{ssb}, employs defect segmentation-like training%
. However, its primary objective remains to enhance classification quality. %
We include it as a baseline to test its applicability to defect pattern segmentation.
To our knowledge, 
no prior work has attempted to
isolate single-defect patterns from mixed-type defects.

\section{Challenges in Defect Pattern Segmentation}

The goal is to separate defect pixels into pattern-specific subsets within a defective region. 
However, %
defect pattern segmentation poses unique challenges that are not present in natural image segmentation.
This is because, although both are image representations, they have basically different characteristics.
We list them as below. 
\begin{enumerate}[leftmargin=*]
    \item \textbf{Difficulty of obtaining high-quality labels.} Obtaining pixel-level %
    annotations is particularly challenging and expensive. 
    The pixel-level labels require substantial efforts from experts, compared to labels for image classification. Moreover,
    common crowdsourcing for annotation is often infeasible in the semiconductor domain due to strict confidentiality requirements. 
    \item \textbf{Binary-valued pixels.} 
    Defect image pixels only have binary values: normal (0) or defective (1).
    This implies that there are limited features to learn from, as the image only comprises a white foreground region with a lack of visual cues.
    Without such cues, 
    models %
    must rely on structural patterns and distributional statistics, 
    making the task uniquely challenging.
    \item \textbf{Sparsity of defect patterns.}
    Defect patterns may exhibit pixel-level sparsity (small defective regions within large backgrounds) and image-level sparsity (rare occurrence of certain patterns).
    As shown in \cref{fig:dataset}, some patterns have extremely sparse appearances, where background (non-defect regions) dominate the image. This indicates pixel-level sparsity. 
    Some patterns may be scarce due to specific causes in manufacturing. This leads to image-level sparsity where data imbalance may occur.

\end{enumerate}
Unlike existing baselines that neglect these factors, \dd explicitly addresses such domain-specific challenges to achieve effective defect pattern segmentation quality.

\section{Methodology}

In \cref{fig:method}, we illustrate the overall training with \dd. 
The entire process operates without expensive pixel-level annotations. 
\dd uses the image-level labels to first learn basic pattern-level features, then leverages this knowledge for identifying key spatial features.
Specifically, \dd trains the model through three phases: 
1) Pretraining the encoder for classification (\cref{sec:method:ph1}), 2) Adapting it to learn spatial cues of defect patterns (\cref{sec:method:ph2}), %
and 3) Final training on synthetic mixed-defect data for precise pattern segmentation (\cref{sec:method:ph3}).
We use the U-Net~\cite{unet} architecture, as it is widely used for segmentation on real-world images.

\subsection{\phaseone: Encoder Training}
\label{sec:method:ph1}
\phaseone initializes the U-Net encoder with basic knowledge required to identify each single-type defect pattern. 
This ability is later exploited to generate cues that guide the model to spatially localize each defect pattern.
In \phaseone, we use the available image-level labels to train the encoder for classification task.
For this, an MLP-based classification head is attached after the encoder to predict the presence of each single-type defect in the image.
The model is trained with the mean of Binary Cross-Entropy Loss for each label, as defect pattern classification is a multi-label task.
\subsection{\phasetwo: Spatial Knowledge Transfer}
\label{sec:method:ph2}

In \phasetwo, we train the full U-Net architecture (encoder and decoder). 
The goal at this stage is to learn spatial indicators of defect patterns.
This is done by training the model with \emph{soft supervision maps} with a pattern-weighted Dice loss.
Given the absence of semantic information in the images, we generate \emph{soft supervision maps} to supply pattern-aware hints, stabilizing the model’s early learning process to differentiate defect patterns. 
The pattern-weighted Dice Loss helps this process by assigning loss weights, ensuring balanced optimization across defect patterns.

\subsubsection{Soft Supervision Maps}
\begin{figure}[b]
\hfill
\centerline{\includegraphics[width=\columnwidth]{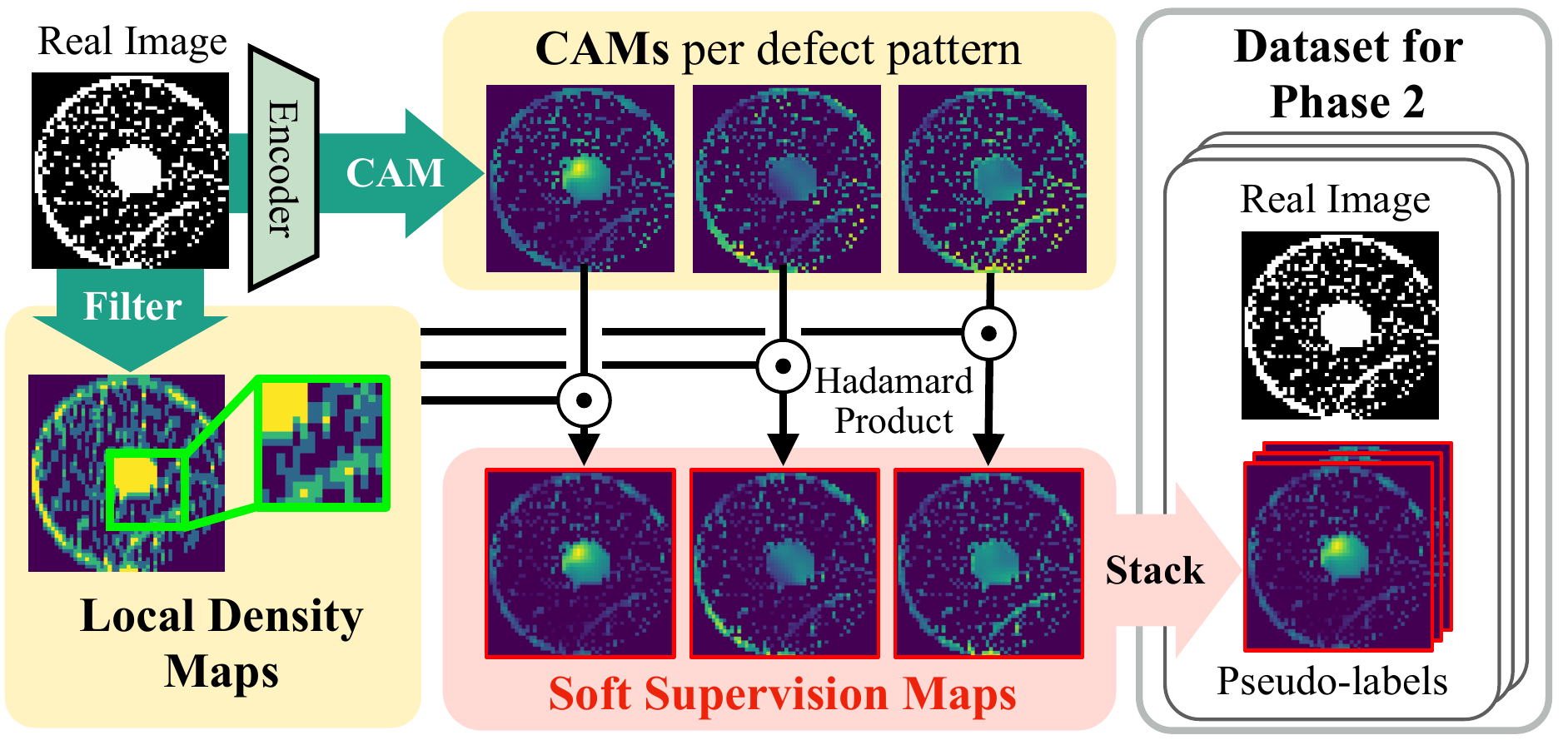}}
\caption{%
Generation of soft supervision maps.
For each image, local density map (bottom left) and CAM (top center) are computed, combined to generate soft supervision maps. %
}
\label{fig:fast_gen}
\end{figure}
Soft supervision maps act as intermediate pseudo-labels that guide the model toward defect regions. 
They incorporate global and local spatial cues, each extracted from the image-level classifier, and the defective regions of the image.
To construct these maps, we combine local density maps and class activation maps (CAM, \cite{gradcam, jacobgilpytorchcam}) from the pretrained classifier.

On the one hand, we generate a local density map of the defect images. 
Local density maps provide %
global cues of the image by highlighting regions where defect pixels are concentrated. 
Defect patterns typically form clusters, %
so we compute these maps using a simple mean filter that amplifies dense defect regions and smooths out sparse areas.
High-value regions in the density map represent probable defect zones, as illustrated in \fig{fig:fast_gen}.

On the other hand, we also utilize CAM, which complements the density maps by identifying regions that influence the classifier’s decision for each defect type. 
While density alone offers spatial importance, it does not differentiate between patterns. 
Therefore, CAM introduces type-specific relevance, enabling the supervision to carry pattern-specific information.

Finally, we generate the final soft supervision maps by multiplying each type-specific CAM with the density map, followed by normalization on pixel values.
These score-like pixel values represent the joint contribution of defect concentration and pattern-specific signals from the classifier. %
The U-Net is trained to align its output with these maps, using the pattern-weighted Dice loss which balances training across defect patterns. %

\subsubsection{Pattern-Weighted Dice Loss}
Due to their sparse and monotonic regions, defect patterns are difficult for the model to separate in the early stages of training. 
This often leads the model to default to predicting mostly absent masks for some patterns. 
To mitigate this issue and encourage the model to first detect coarse regions of patterns, thereby improving early-stage recall, we introduce the pattern-weighted Dice Loss.
We use the loss to ensure the model to primarily capture the existing defects and further enable balanced training for all defect patterns, regardless of the available samples of each defect pattern. 

The original Dice loss~\cite{diceloss} is widely used for segmentation tasks:
\begin{align}
\mathcal{L}_{\mathrm{dice},i,p} = 1 -
\frac{2  (\sum y_{i,p} \cdot \hat{y}_{i,p}) + \epsilon}
{ \sum y_{i,p} + \sum \hat{y}_{i,p} + \epsilon}
\label{eq:dice}
\end{align}
where $\hat{y}_{i,p}$ denotes the predicted mask and $y_{i,p}$ denotes the ground truth mask for image $i$ and single-type pattern $p$. 
We observed that directly using the Dice loss led to a drastic degradation of less frequent and scarce defect patterns.
This behavior stems from the loss, pushing the model toward absent-mask predictions for rarely occurring patterns.
Therefore, we assign loss weights to every image-pattern pair to help the model focus more on the presence of each pattern, rather than being biased by the absence of patterns.
For each image $i$ and pattern $p$, we define the indicator:
\begin{equation}
\text{Let } \delta_{i,p} =
\begin{cases}
1, & \text{if pattern } p \text{ exists in the image } i, \\
0, & \text{otherwise.}
\end{cases}
\end{equation}

\noindent
Let the dataset-level occurrence count of pattern $p$ be
$K_p = \sum_{i=1}^{N} \delta_{i,p}$
where $N$ is the total number of images in the train set.  

The Pattern-Weighted Dice Loss is then defined as

\begin{equation}
\mathcal{L}_{\text{PWDL}}
=
\sum_{p=1}^{P}
\frac{1}{K_p}
\sum_{i=1}^{N}
\delta_{i,p}
\left(
1 -
\frac{2 (\sum y_{i,p} \cdot \hat{y}_{i,p}) + \epsilon}
{ \sum y_{i,p} + \sum \hat{y}_{i,p} + \epsilon}
\right),
\label{eq:pwdl}
\end{equation}
\noindent
Equivalently,
\begin{equation}
\mathcal{L}_{\text{PWDL}}
=
\sum_{p=1}^{P}
\frac{1}{K_p + \epsilon}
\sum_{i : \delta_{i,p}=1}
\mathcal{L}_{\mathrm{dice},i,p} 
\end{equation}

The Pattern-Weighted Dice Loss %
stabilizes learning across patterns of different prevalence and improves segmentation on scarce defect patterns.

\subsection{\phasethree: Synthetic Defect Training}
\label{sec:method:ph3}
\phasethree further refines the performance of \dd with fine-tuning using a synthetic dataset with accurate segmentation masks.
\subsubsection{Synthetic Dataset Generation}
\begin{figure}[b]
    \centering
    \includegraphics[width=0.98\linewidth]{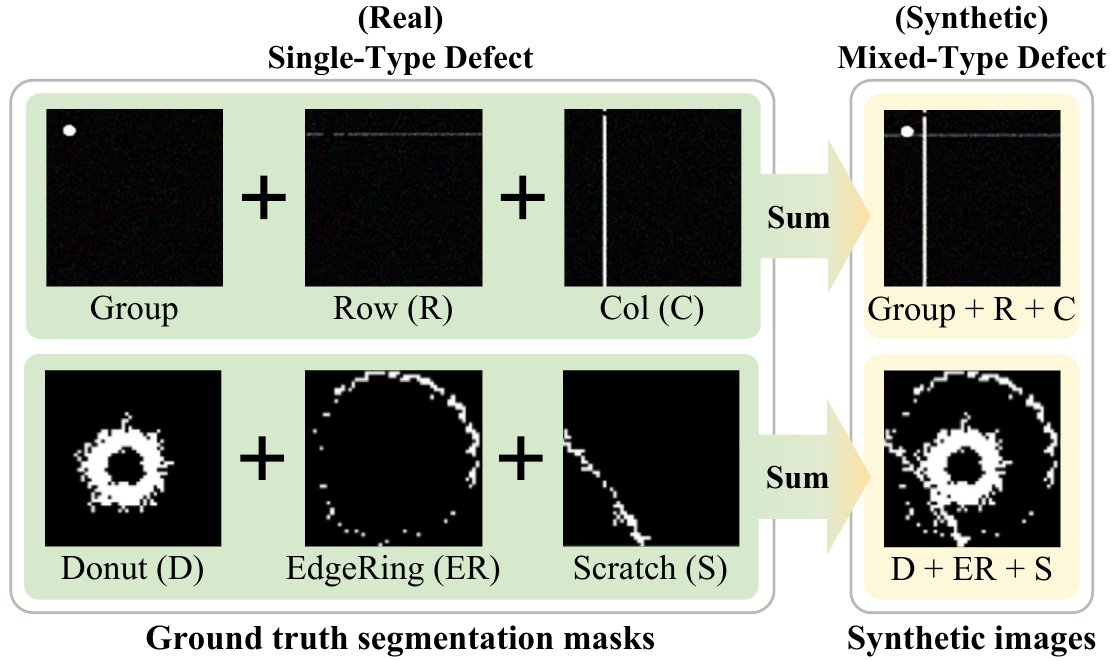}
    \caption{Generation of synthetic mixed-type defect images. %
    }
    \label{fig:synthetic}
\end{figure}

As discussed in the previous section, because the datasets are binary-valued images, any 1-valued pixels correspond to one of the target patterns.
Therefore, for single-type defects, all 1-valued pixels can directly serve as the segmentation masks for those patterns.
The remaining problem is that this only applies to the single-type defects, and not to the mixed-type defects.
To tackle this problem, we 
synthesize multi-type defect masks by merging single-type defect masks.
For example, in the \mtdata dataset, a row-defect map and a column-defect map can simply be added together with slight shifts for augmentation. 
This creates a synthetic, but realistic mixed-type defect maps,
which are shown in Figure \ref{fig:synthetic}.
Thus, for the training, we use the synthetic dataset's segmentation masks as the ground truth.

\input{tabs/loc_eval_ap}

\subsubsection{Absence Penalty Loss \& Scheduling}
When training the model in \phasethree, we train the model with the aforementioned Pattern-Weighted Dice Loss and the Absence Penalty (AP) Loss. 
Here, the Absence Penalty Loss is introduced to reduce the false-positive pixels. 
Since the absent sample pairs are ignored in \phasetwo by design~\Cref{eq:pwdl},
the trained model has a tendency to include irrelevant pixels, or sometimes show defect patterns that are not present. 
To address this, we add a loss term that penalizes such a prediction with Binary Cross-Entropy (BCE) Loss. 
\begin{equation}
\text{Let}\   
 \hat{y'}_{i,p}(x) =
\begin{cases}
\hat{y}_{i,p}(x), & \text{if} \ \hat{y}_{i,p}(x) \ge \tau \\
0, & \text{otherwise.}
\end{cases}
\end{equation}

\noindent
where $x$ indexes the location of pixels in $\hat{y}_{i,p}$, and $\tau$ is the confidence threshold. 
If the confidence is greater than or equal to $\tau$, it is included in the loss. 
We gradually reduce the value of $\tau$ from 0.5 to 0.0, depending on loss convergence.
Define the pixel-wise BCE against an all-zero target as
\begin{equation}
\mathcal{L}_\mathrm{BCE}\!\left(\hat{y'}_{i,p}, 0\right)
= -\frac{1}{|\Omega|}
\sum_{x \in \Omega}
\log\!\big(1 - \hat{y'}_{i,p}(x) + \epsilon\big),
\end{equation}
where $\Omega$ denotes the set of all pixels above the threshold in the image for $\hat{y'}_{i, p}$.
\noindent
Then the Absence Penalty loss is
\begin{equation}
\mathcal{L}_{\text{AP}} =
\sum_{p=1}^{P}
\frac{1}{K_p'+ \epsilon}
\sum_{i=1}^{N}
a_{i,p}\,\mathcal{L}_\mathrm{BCE}\!\left(\hat{y'}_{i,p}, 0\right),
\end{equation}
where $a_{i,p} = 1-\delta_{i,p}$ and $\quad K_p' = \sum_{i=1}^{N} 1-\delta_{i,p}$.

\noindent
Consequently, the total loss is
\begin{equation}
\mathcal{L}_{\text{total}}
= \lambda_{\text{dice}}\,\mathcal{L}_{\text{PWDL}}
+ \lambda_{\text{ap}}\,\mathcal{L}_{\text{AP}},
\end{equation}
where $\lambda_{\text{dice}}, \lambda_{\text{ap}} \ge 0$ are weighting hyperparameters.

\section{Experimental Results}
\label{sec:experimental}

\subsection{Experimental Settings}
We use two datasets for evaluation: \mtdata\ and \wmdata. Each dataset was randomly split into 80\% training, 10\% validation, and from the remaining 10\%, we manually annotated 1,000 samples and used them as test sets. %
Model prediction was performed for defective regions in each image.
In \wmdata, %
all non-defective pixels are unified to 0, resulting in binary images.
While class labels were provided, test-set masks were manually annotated at the pixel level.
Our framework was implemented with PyTorch \cite{pytorch}. 
Experiments were run on NVIDIA RTX 4090 GPUs.

We chose baselines %
compatible with our scenario: Segmentation methods that do not require pixel-level annotations (See \ref{sec:wsss}):
CLIPSeg \cite{clipseg}, MCT \cite{mctformer}, ReCAM \cite{recam},  S2C \cite{sam2cams} and SSB-Rec \cite{ssb}.
CLIPSeg \cite{clipseg} is a widely used zero-shot segmentation model. 
S2C \cite{sam2cams} generates pixel-level masks refined by SAM \cite{SAM}. 
MCT \cite{mctformer} uses class-specific tokens in the transformer to obtain discriminative attention maps.
ReCAM \cite{recam} introduces a loss function to improve object boundaries in CAM-based pseudo-labels.
Lastly, SSB-Rec \cite{ssb} also leverages the idea of generating a pseudo dataset with single-type defect images, but uses it to train classification models. 
We modify the method to do a segmentation task, which is trivially done as it already uses U-Net with a segmentation-based loss.

\subsection{Comparison of Defect Segmentation Results}
\subsubsection{Performance Comparison}
\Cref{tabs/loc_eval_ap} shows the performance comparison by evaluating mean Intersection-over-Union (mIoU) results.
As shown in \cref{tabs/loc_eval_ap}, \dd outperforms baselines in both datasets by a great margin.
For the \mtdata dataset, the average mIoU is 67.42\%, which shows a superior value to be used in practice. 
For the \wmdata dataset, the average mIoU is 69.04\%, also showing a reliable performance, considering the larger number of classes.
On the other hand, the baselines do not show practical performance, or even fail to work on some patterns.
This degradation occurs because baselines rely on rich visual features that are not present in binary defect images. 
Therefore, its general knowledge does not help the segmentation of the defect patterns when using only weak visual cues from the binary pixels.
Unlike the baselines, 
only \dd successfully performs defect pattern segmentation, validating the design and its suitability for practical use.

\begin{figure}[b]
\centerline{\includegraphics[width=\columnwidth]{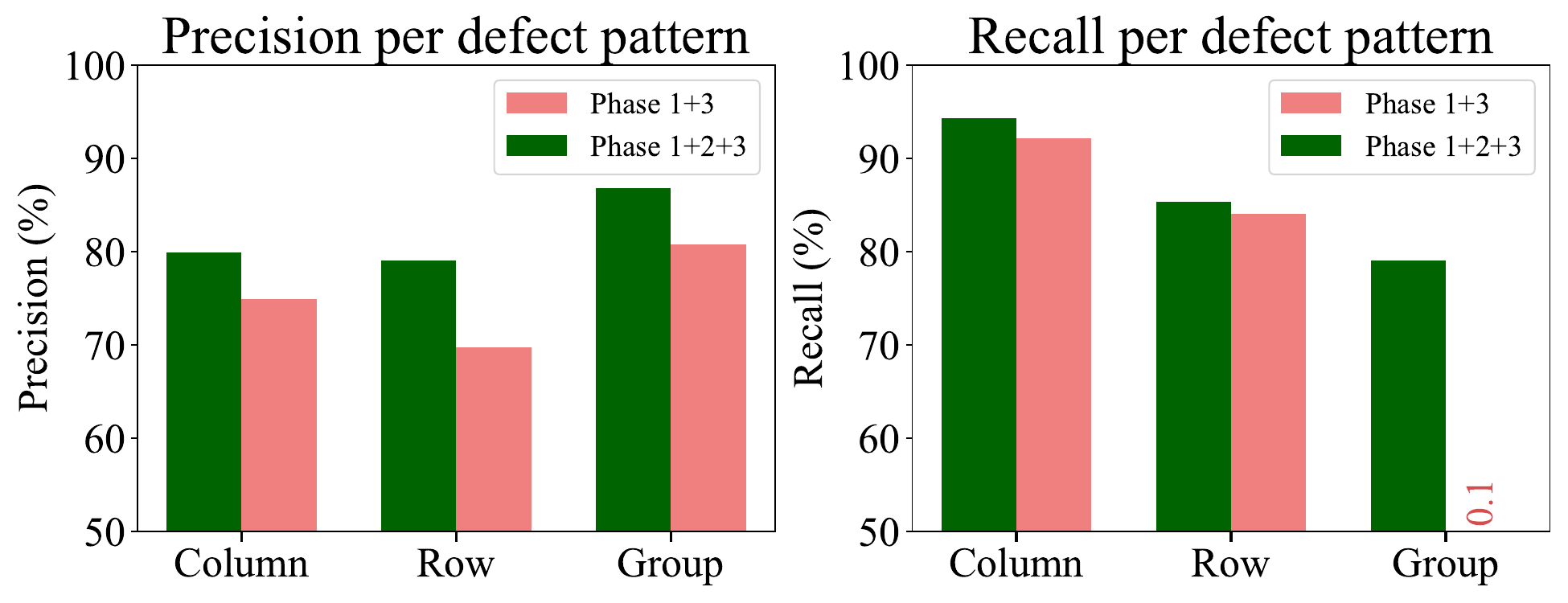}}
\caption{Precision and Recall comparison with and without \phasetwo in \dd, with \mtdata. 
It shows improved metrics when trained using all 3 phases.
}
\label{loss_curve}
\end{figure}
\subsubsection{Visualizations}
\begin{figure*}[ht]
    \centering
    \includegraphics[width=0.9\textwidth]{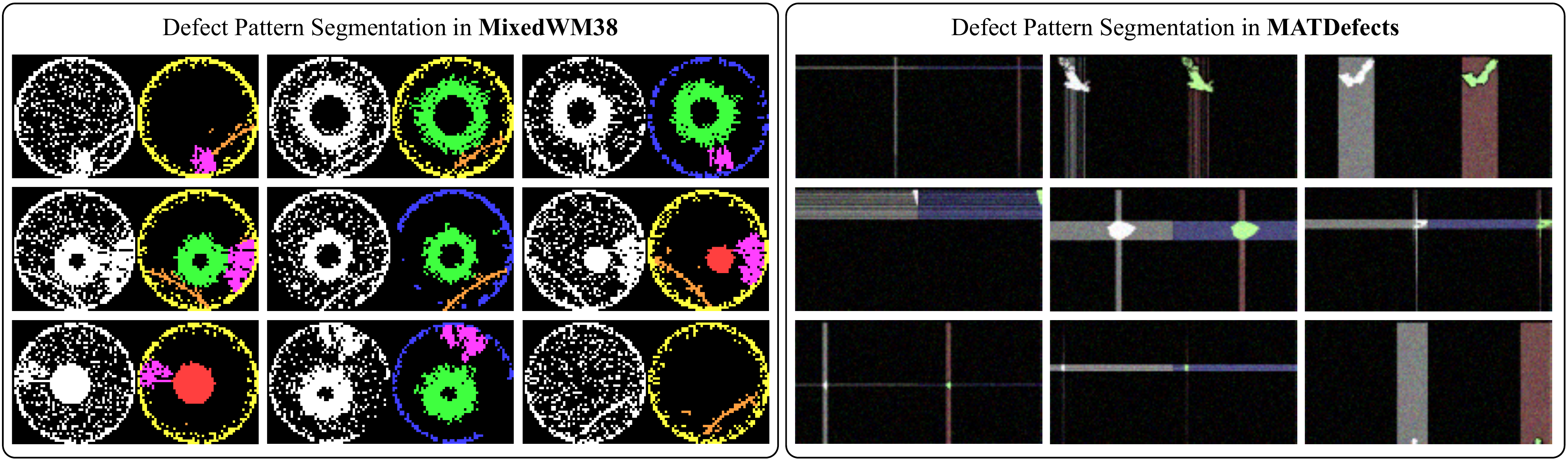}
    \caption{Examples of segmented defect pattern in test images, using our proposed method, \dd. For proprietary reasons, the examples of \mtdata\ were slightly modified from the original data. Key %
    patterns were preserved regardless of modification.}
    \label{fig:vis_loc}
\end{figure*}
We visualize defect segmentation results for the test images of \mtdata and \wmdata in \fig{fig:vis_loc}. 
Each example shows the original defect image where multiple pattern types overlap. %
\dd successfully disentangles these mixed patterns.  %
Pixels of the same pattern share identical colors.
The segmented regions exhibit clear morphological distinctions that accurately represent each defect pattern.
This shows the practical value of leveraging \dd for automated defect analysis, preserving critical pattern and shape information.

\begin{table}[]
\caption{Ablation Study of \dd with \mtdata}
\centering
\resizebox{\columnwidth}{!}{%
\begin{tabular}{ccc|cccc}
\toprule
\multirow{2}{*}[-0.3em]{\phaseone}&\multirow{2}{*}[-0.3em]{\phasetwo}&\multirow{2}{*}[-0.3em]{\phasethree}&\multicolumn{4}{c}{mIoU (\%)} \\
\cmidrule(lr){4-7}
&&&Column&Row&Group&\textbf{Mean} \\
\midrule
\cmark&\xmark&\xmark&0.00&0.00&0.61&0.20\\
\cmark&\cmark&\xmark&42.18&42.32&33.98&39.49\\
\cmark&\xmark&\cmark&62.30&56.54&0.14&39.66\\ %
\midrule
\rowcolor{blue!10}
\cmark&\cmark&\cmark&\textbf{70.91}&\textbf{66.66}&\textbf{64.68}&\textbf{67.42}\\
\bottomrule
\end{tabular}
}
\label{tab:ablation}
\end{table}

\subsection{Further Analyses of \dd}
\begin{figure}[b]
\centerline{\includegraphics[width=0.87\columnwidth]{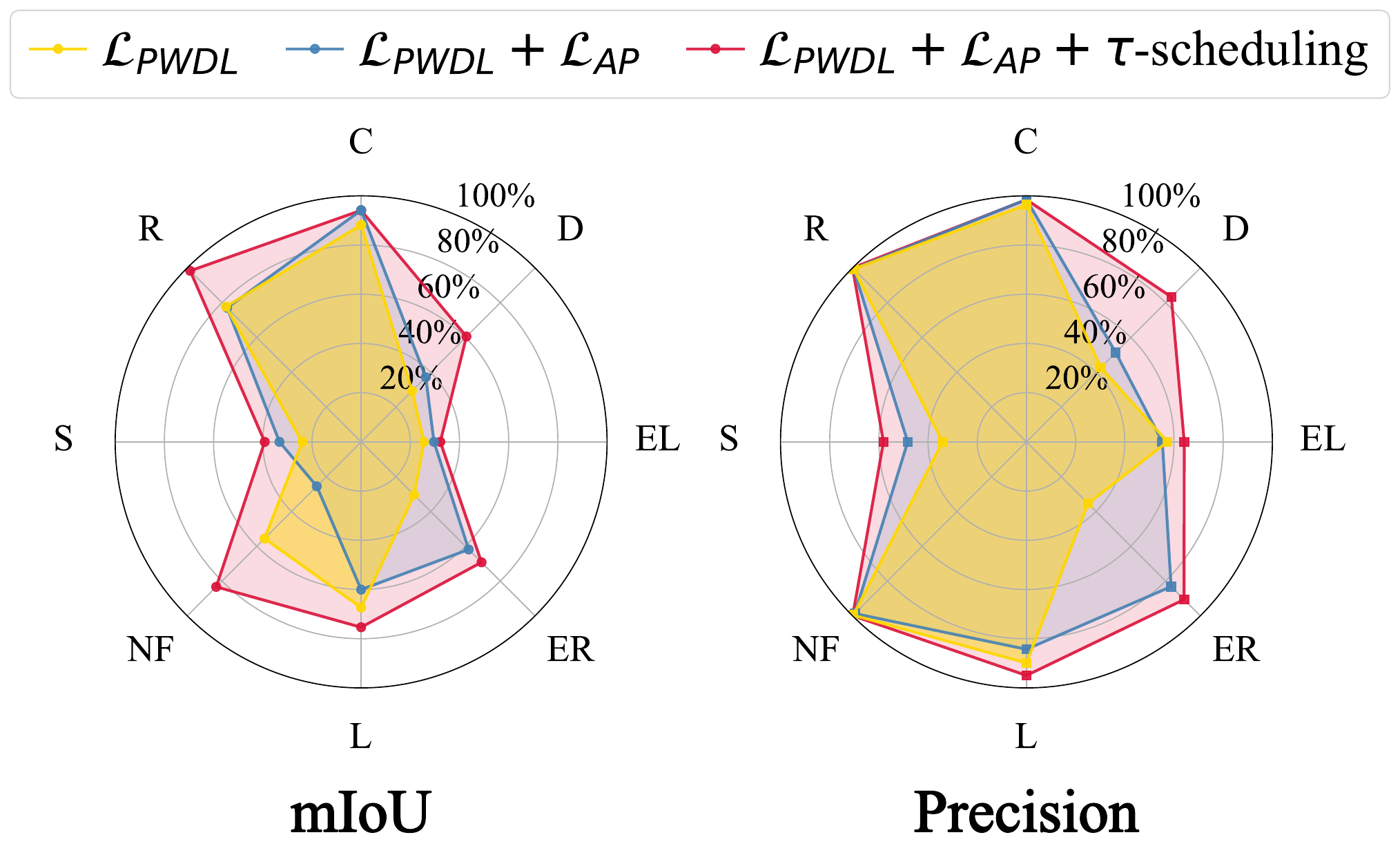}}
\caption{Loss term ablation for \phasethree with \wmdata. The left figure compares mIoU across single-type defect patterns, and the
right depicts the difference in Precision.}
\label{radar}
\end{figure}
\subsubsection{Ablation Study of Phases}

\tab{tab:ablation} shows the ablation study of \dd %
where we evaluate
models trained with selective phases.

The first row shows that \phaseone yields almost no segmentation capability, with a mIoU of 0.20\%. 
When adding \phasetwo, the mean score increases to 39.49\%, indicating that soft supervision helps the model learn spatial cues but remains insufficient. 
Similarly, using only \phasethree with \phaseone increases the mean mIoU further to 39.66\%, showing that refinement is effective but still incomplete, as the model fails on "group" patterns. 
The final setting with all phases corresponds to \dd, achieving the highest performance with a mean mIoU of 67.42\%, validating the overall design.

A notable observation is the substantial mIoU gap in "group" pattern segmentation when comparing the third and final rows.
Without soft supervision maps (third row), the model %
fails to isolate group defects. %
This confirms the crucial role of Phase 2. %
As group defects typically co-occur with line-defects (\fig
{fig:vis_loc}), \phasetwo proves essential for 
disentangling such overlapping patterns. %

We further analyze the synergistic effects of phases in \dd, %
shown in \fig{loss_curve} with \mtdata.
The results for the model trained with all phases of \dd are shown in green.
The results in red are obtained only with \phaseone and \phasethree. 
Notably, the 0.1\% recall for the "group" pattern when \phasetwo is excluded suggests that the model 
only captures a very small portion of true defect pixels.
The model collapses without an early-stage understanding of spatial context about defect patterns, highlighting \phasetwo's primary role in supplying knowledge of defect regions. %
By comparing the two, 
we can see that the overall segmentation is better with \phasetwo.%

\subsubsection{Loss Term Analysis}
\fig{radar} illustrates the %
improvement in segmentation as additional loss terms and scheduling strategies are added to \phasethree.
From left to right, each setting introduces a new component: \emph{Pattern-Weighted Dice Loss ($\mathcal{L}_{PWDL}$) only}, \emph{$\mathcal{L}_{PWDL}$ + Absence Penalty ($\mathcal{L}_{AP}$)}, and finally \emph{$\mathcal{L}_{PWDL}$ + $\mathcal{L}_{AP}$ + $\tau$-Scheduling}.
This %
results in a gradual increase in IoU across all defect patterns.

\emph{Pattern-Weighted Dice Loss Only} serves as the baseline. %
While it handles class imbalance, it shows limited performance %
on sparse patterns (e.g., \textit{Scratch}, \textit{EdgeRing}), which remain underrepresented. %
Adding the \emph{Absence Penalty} term provides negative supervision for pattern-absent regions, reducing false positives.
This yields noticeable IoU gains on previously over-predicted patterns 
(e.g., \textit{EdgeRing}). %
Finally, \emph{$\tau$-Scheduling} further improves the model by suppressing all false-positive pixels during training. 
As training progresses, decreasing $\tau$ makes the model more conservative, reducing false positives and improving IoU for nearly all patterns.
Overall, the figure demonstrates the cumulative benefit of the three phases, 
showing the best performance when used altogether.

\section{Conclusion}
In this work, we present \dd, a pioneering weakly supervised framework that advances automatic semiconductor defect inspection from image-level classification to pattern-level segmentation. 
Unlike conventional approaches requiring costly pixel-level annotations, our method uses only image-level labels to separate defect instances, %
while remaining scalable and cost-effective. 
Experiments on defect datasets show that \dd\ consistently outperforms baselines.
By providing a single type of pattern-level feedback for defects, our framework can enhance optical inspection, long-term monitoring, and continuous manufacturing improvement.

\begin{acks}
This paper was the result of the research project supported by SK Hynix Inc (MDUS001T).
Jinho Lee is the corresponding author.
\end{acks}

\newpage

\bibliographystyle{ACM-Reference-Format}
\bibliography{cite}

\end{document}

%% file: packages.tex
\usepackage{comment}
\usepackage{booktabs}
\usepackage{multicol}
\usepackage{multirow}
\usepackage{cleveref}
\crefname{figure}{Figure}{Figures}
\Crefname{figure}{Figure}{Figures}
\usepackage{enumitem}
\usepackage{xspace}
\usepackage[table]{xcolor}
\definecolor{olivegreen}{rgb}{0, 0.6, 0}

\newcommand{\fig}[1]{Figure~\ref{#1}}
\newcommand{\tab}[1]{Table~\ref{#1}}

\newcommand{\dd}{SePArate\xspace}
\newcommand{\mtdata}{MATDefects\xspace}
\newcommand{\wmdata}{MixedWM38\xspace}

\newcommand{\phaseone}{Phase 1\xspace}
\newcommand{\phasetwo}{Phase 2\xspace}
\newcommand{\phasethree}{Phase 3\xspace}

\usepackage{pifont}%
\newcommand{\cmark}{\color{olivegreen}\ding{51}}%
\newcommand{\xmark}{\color{red}\ding{55}}%

%% file: tabs/loc_eval_ap.tex
\begin{table*}[ht]
\centering
\caption{Comparison of Mean IoU (mIoU) for Defect Pattern Segmentation (\%)}
\resizebox{\textwidth}{!}{%
\begin{tabular}{l|cccc|ccccccccc}

\toprule
\multirow{2}{*}[-0.4em]{\centering Method} 
      & \multicolumn{4}{c|}{\mtdata} 
      & \multicolumn{9}{c}{\wmdata} \\
\cmidrule(lr){2-5}\cmidrule(lr){6-14}
&Column&Row&Group&\textbf{Mean} &Center&Donut&EdgeLoc&EdgeRing&Loc&NearFull&Scratch&Random& \textbf{Mean} \\
\midrule
CLIPSeg~\cite{clipseg}&4.20&4.32&4.12&4.21&5.16&5.06&4.94&4.82&4.87&4.87&5.17&5.28&5.02\\
MCT~\cite{mctformer} %
&42.53&50.60&31.66&41.60&5.60&16.89&15.34&33.73&5.74&21.43&4.36&82.86&23.24\\
ReCAM~\cite{recam} &37.23&0.00&23.71&20.31&10.52&37.51&9.63&11.56&8.69&63.92&8.33&93.55&30.46\\
S2C~\cite{sam2cams} &60.57&40.13&32.49&44.40&1.43&0.20&13.20&10.75&18.18&0.00&8.90&0.00&6.58\\
SSB-Rec~\cite{ssb}&55.59&50.89&44.81&50.43&5.44&18.71&0.84&26.57&56.32&32.94&16.43&67.53&28.10\\

\midrule
\rowcolor{blue!10}
Ours, \textbf{\dd} &70.91&66.66&64.68&\textbf{67.42}&94.10&60.56&32.20&69.23&75.35&83.33&39.21&98.31&\textbf{69.04}\\
\bottomrule
\multicolumn{14}{l}{$^*$ MCT: MCTFormer, SSB-Rec: Semantic Segmentation-Based Wafer Map
Mixed-Type Defect Pattern Recognition}
\end{tabular}
}
\label{tabs/loc_eval_ap}
\end{table*}

%% file: cite.bib
@String{Computing = "Computing" }

@String{Computer = "{IEEE} Computer" }

@String{Springer = "Springer-Verlag" }

@article{mixedwm38,
  title={Deformable convolutional networks for efficient mixed-type wafer defect pattern recognition},
  author={Wang, Junliang and Xu, Chuqiao and Yang, Zhengliang and Zhang, Jie and Li, Xiaoou},
  journal={IEEE Transactions on Semiconductor Manufacturing},
  volume={33},
  number={4},
  pages={587--596},
  year={2020},
  publisher={IEEE}
}

@inproceedings{palm,
  title={PaLM: Point Cloud and Large Pre-trained Model Catch Mixed-type Wafer Defect Pattern Recognition},
  author={He, Hongquan and Kuang, Guowen and Sun, Qi and Geng, Hao},
  booktitle={2024 Design, Automation \& Test in Europe Conference \& Exhibition (DATE)},
  pages={1--6},
  year={2024},
  organization={IEEE}
}

@inproceedings{waferssm,
  title={Efficient Modulated State Space Model for Mixed-Type Wafer Defect Pattern Recognition},
  author={Nie, Mu and Zhu, Shidong and Yan, Aibin and Zhuo, Cheng and Wen, Xiaoqing and Ni, Tianming},
  booktitle={2025 Design, Automation \& Test in Europe Conference (DATE)},
  pages={1--6},
  year={2025},
  organization={IEEE}
}

@article{shankar2005defect,
  title={Defect detection on semiconductor wafer surfaces},
  author={Shankar, NG and Zhong, ZW},
  journal={Microelectronic engineering},
  volume={77},
  number={3-4},
  pages={337--346},
  year={2005},
  publisher={Elsevier}
}

@inproceedings{gleason1998rapid,
  title={Rapid yield learning through optical defect and electrical test analysis},
  author={Gleason, Shaun S and Tobin Jr, Kenneth W and Karnowski, Thomas P and Lakhani, Fred},
  booktitle={Metrology, Inspection, and Process Control for Microlithography XII},
  volume={3332},
  pages={232--242},
  year={1998},
  organization={SPIE}
}

@article{getachew2024data,
  title={Data analytics in zero defect manufacturing: a systematic literature review and proposed framework},
  author={Getachew, Mehret and Beshah, Birhanu and Mulugeta, Ameha},
  journal={International Journal of Production Research},
  pages={1--33},
  year={2024},
  publisher={Taylor \& Francis}
}

@article{pytorch,
  title={Pytorch: An imperative style, high-performance deep learning library},
  author={Paszke, Adam and Gross, Sam and Massa, Francisco and Lerer, Adam and Bradbury, James and Chanan, Gregory and Killeen, Trevor and Lin, Zeming and Gimelshein, Natalia and Antiga, Luca and others},
  journal={Advances in neural information processing systems},
  volume={32},
  year={2019}
}

@misc{waferdata1,
    title={Modernizing Semiconductor Manufacturing Analytics Platform using AWS Data Analytics Services},
author={Upasana Pandya},
howpublished = {\url{https://designthesolution.org/wp-content/uploads/2024/10/process-track-upasana-modernizing-semiconductor-manufacturing-analytics-aws-data-analytics-services.pdf}}
}

@misc{waferdata2,
    title={Too Much Fab And Test Data, Low Utilization
},
author={Anne Meixner},
howpublished = {\url{https://semiengineering.com/too-much-fab-and-test-data-low-utilization/}}
}

@INPROCEEDINGS{isscc_ddr4,
  author={Koo, Kibong and Ok, Sunghwa and Kang, Yonggu and Kim, Seungbong and Song, Choungki and Lee, Hyeyoung and Kim, Hyungsoo and Kim, Yongmi and Lee, Jeonghun and Oak, Seunghan and Lee, Yosep and Lee, Jungyu and Lee, Joongho and Lee, Hyungyu and Jang, Jaemin and Jung, Jongho and Choi, Byeongchan and Kim, Yongju and Hur, Youngdo and Kim, Yunsaing and Chung, Byongtae and Kim, Yongtak},
  booktitle={IEEE International Solid-State Circuits Conference}, 
  title={A 1.2V 38nm 2.4Gb/s/pin 2Gb DDR4 SDRAM with bank group and ×4 half-page architecture}, 
  year={2012}
}

@inproceedings{SAM,
  title={Segment anything},
  author={Kirillov, Alexander and Mintun, Eric and Ravi, Nikhila and Mao, Hanzi and Rolland, Chloe and Gustafson, Laura and Xiao, Tete and Whitehead, Spencer and Berg, Alexander C and Lo, Wan-Yen and others},
  booktitle={Proceedings of the IEEE/CVF international conference on computer vision},
  pages={4015--4026},
  year={2023}
}

@inproceedings{clipseg,
  title={Image segmentation using text and image prompts},
  author={L{\"u}ddecke, Timo and Ecker, Alexander},
  booktitle={Proceedings of the IEEE/CVF conference on computer vision and pattern recognition},
  pages={7086--7096},
  year={2022}
}

@inproceedings{gradcam,
  title={Grad-cam: Visual explanations from deep networks via gradient-based localization},
  author={Selvaraju, Ramprasaath R and Cogswell, Michael and Das, Abhishek and Vedantam, Ramakrishna and Parikh, Devi and Batra, Dhruv},
  booktitle={Proceedings of the IEEE international conference on computer vision},
  pages={618--626},
  year={2017}
}

@inproceedings{diceloss,
  title={V-net: Fully convolutional neural networks for volumetric medical image segmentation},
  author={Milletari, Fausto and Navab, Nassir and Ahmadi, Seyed-Ahmad},
  booktitle={2016 fourth international conference on 3D vision (3DV)},
  pages={565--571},
  year={2016},
  organization={Ieee}
}

@inproceedings{unet,
  title={U-net: Convolutional networks for biomedical image segmentation},
  author={Ronneberger, Olaf and Fischer, Philipp and Brox, Thomas},
  booktitle={International Conference on Medical image computing and computer-assisted intervention},
  pages={234--241},
  year={2015},
  organization={Springer}
}

@article{ssb,
  title={Semantic segmentation-based wafer map mixed-type defect pattern recognition},
  author={Yan, Jinda and Sheng, Yi and Piao, Minghao},
  journal={IEEE Transactions on Computer-Aided Design of Integrated Circuits and Systems},
  volume={42},
  number={11},
  pages={4065--4074},
  year={2023},
  publisher={IEEE}
}

@inproceedings{recam,
  title={Class re-activation maps for weakly-supervised semantic segmentation},
  author={Chen, Zhaozheng and Wang, Tan and Wu, Xiongwei and Hua, Xian-Sheng and Zhang, Hanwang and Sun, Qianru},
  booktitle={Proceedings of the IEEE/CVF conference on computer vision and pattern recognition},
  pages={969--978},
  year={2022}
}

@inproceedings{sam2cams,
  title={From sam to cams: Exploring segment anything model for weakly supervised semantic segmentation},
  author={Kweon, Hyeokjun and Yoon, Kuk-Jin},
  booktitle={Proceedings of the IEEE/CVF Conference on Computer Vision and Pattern Recognition},
  pages={19499--19509},
  year={2024}
}

@inproceedings{mctformer,
  title={Multi-class token transformer for weakly supervised semantic segmentation},
  author={Xu, Lian and Ouyang, Wanli and Bennamoun, Mohammed and Boussaid, Farid and Xu, Dan},
  booktitle={Proceedings of the IEEE/CVF conference on computer vision and pattern recognition},
  pages={4310--4319},
  year={2022}
}

@article{wafersegnet, 
  title={WaferSegClassNet-A light-weight network for classification and segmentation of semiconductor wafer defects},
  author={Nag, Subhrajit and Makwana, Dhruv and Mittal, Sparsh and Mohan, C Krishna and others},
  journal={Computers in Industry},
  volume={142},
  pages={103720},
  year={2022},
  publisher={Elsevier}
}

@article{nakazawa2019anomaly,
  title={Anomaly detection and segmentation for wafer defect patterns using deep convolutional encoder--decoder neural network architectures in semiconductor manufacturing},
  author={Nakazawa, Takeshi and Kulkarni, Deepak V},
  journal={IEEE Transactions on Semiconductor Manufacturing},
  volume={32},
  number={2},
  pages={250--256},
  year={2019},
  publisher={IEEE}
}

@article{groundedsam,
  title={Grounded sam: Assembling open-world models for diverse visual tasks},
  author={Ren, Tianhe and Liu, Shilong and Zeng, Ailing and Lin, Jing and Li, Kunchang and Cao, He and Chen, Jiayu and Huang, Xinyu and Chen, Yukang and Yan, Feng and others},
  journal={arXiv preprint arXiv:2401.14159},
  year={2024}
}

@article{kyeong2018classification,
  title={Classification of mixed-type defect patterns in wafer bin maps using convolutional neural networks},
  author={Kyeong, Kiryong and Kim, Heeyoung},
  journal={IEEE Transactions on Semiconductor Manufacturing},
  volume={31},
  number={3},
  pages={395--402},
  year={2018},
  publisher={IEEE}
}

@article{kim2023advances,
  title={Advances in machine learning and deep learning applications towards wafer map defect recognition and classification: a review},
  author={Kim, Tongwha and Behdinan, Kamran},
  journal={Journal of Intelligent Manufacturing},
  volume={34},
  number={8},
  pages={3215--3247},
  year={2023},
  publisher={Springer}
}

@article{yu2019enhanced,
  title={Enhanced stacked denoising autoencoder-based feature learning for recognition of wafer map defects},
  author={Yu, Jianbo},
  journal={IEEE Transactions on Semiconductor Manufacturing},
  volume={32},
  number={4},
  pages={613--624},
  year={2019},
  publisher={IEEE}
}

@article{shon2021unsupervised,
  title={Unsupervised pre-training of imbalanced data for identification of wafer map defect patterns},
  author={Shon, Ho Sun and Batbaatar, Erdenebileg and Cho, Wan-Sup and Choi, Seong Gon},
  journal={IEEE Access},
  volume={9},
  pages={52352--52363},
  year={2021},
  publisher={IEEE}
}

@article{saqlain2020deep,
  title={A deep convolutional neural network for wafer defect identification on an imbalanced dataset in semiconductor manufacturing processes},
  author={Saqlain, Muhammad and Abbas, Qasim and Lee, Jong Yun},
  journal={IEEE Transactions on Semiconductor Manufacturing},
  volume={33},
  number={3},
  pages={436--444},
  year={2020},
  publisher={IEEE}
}

@misc{jacobgilpytorchcam,
  title={PyTorch library for CAM methods},
  author={Jacob Gildenblat and contributors},
  year={2021},
  publisher={GitHub},
  howpublished={\url{https://github.com/jacobgil/pytorch-grad-cam}},
}
